%% file: main_final.tex
\documentclass[10pt,twocolumn,letterpaper]{article}

\usepackage[colorlinks,bookmarks=false]{hyperref}
\usepackage{cvpr}
\usepackage{mmstyle}
\usepackage{times}
\usepackage{epsfig}
\usepackage{graphicx}
\usepackage{amssymb}

\usepackage{enumitem}
\usepackage{subcaption}
\usepackage{booktabs}
\usepackage{multirow}

\cvprfinalcopy % *** Uncomment this line for the final submission

\def\cvprPaperID{1145} % *** Enter the CVPR Paper ID here

\ifcvprfinal\fi
\begin{document}

%%%%%%%%% TITLE
\title{Discover and Learn New Objects from Documentaries}

\author{Kai Chen \hspace{9pt} Hang Song \hspace{9pt} Chen Change Loy \hspace{9pt} Dahua Lin\\
Department of Information Engineering, The Chinese University of Hong Kong \\
{\tt\small ck015,hsong,ccloy,dhlin@ie.cuhk.edu.hk}
% For a paper whose authors are all at the same institution,
% omit the following lines up until the closing ``}''.
% Additional authors and addresses can be added with ``\and'',
% just like the second author.
% To save space, use either the email address or home page, not both
}

\maketitle
\thispagestyle{empty}

%%%%%%%%% ABSTRACT
\begin{abstract}
   \input{abstract.tex}
\end{abstract}

%%%%%%%%% BODY TEXT

\input{introduction.tex}

\input{related-work.tex}

\input{dataset.tex}

\input{framework.tex}
\input{bootstrap.tex}
\input{model.tex}

\input{experiments.tex}
\input{conclusion.tex}

%\clearpage

{\small
\bibliographystyle{ieee}
\bibliography{egbib.bib}
}

\end{document}

%% file: abstract.tex
Despite the remarkable progress in recent years, detecting objects in a new
context remains a challenging task. Detectors learned from a public dataset can
only work with a fixed list of categories, while training from scratch usually
requires a large amount of training data with detailed annotations. This work aims
to explore a novel approach -- learning object detectors from documentary films in
a weakly supervised manner. This is inspired by the observation that documentaries
often provide dedicated exposition of certain object categories, where visual
presentations are aligned with subtitles. We believe that object detectors can be
learned from such a rich source of information. Towards this goal, we develop a
joint probabilistic framework, where individual pieces of information, including
video frames and subtitles, are brought together via both visual and linguistic
links. On top of this formulation, we further derive a weakly supervised learning
algorithm, where object model learning and training set mining are unified in an
optimization procedure. Experimental results on a real world dataset demonstrate
that this is an effective approach to learning new object detectors.

%% file: introduction.tex
\vspace{-0.5cm}
\section{Introduction}
\label{sec:intro}

Recent years have witnessed a wave of innovation in object detection 
driven by the advances in deep learning \cite{girshick2015fast,ouyang2015deepid,ren2015faster}. 
%On ILSVRC, a large-scale benchmark on ImageNet, 
%the mean AP of the object detection task has increased from $22.6\%$ to $66.3\%$ 
%in just four years (from year 2013 to 2016). 
Despite the great success reported on public benchmarks, 
practical applications of such techniques are impeded by 
a significant obstacle, namely, the lack of annotated data. 
Specifically, detectors pre-trained on public datasets 
can only cover a limited list of object categories, 
which are often not sufficient for real-world applications, 
where the objects of interest can go beyond such lists. 
On the other hand, training a new detector requires 
a large quantity of annotated images with bounding boxes provided for individual objects. 
Obtaining such a dataset is an investment that takes tremendous amount of time and resources. 

\begin{figure}[t]
\begin{center}
\includegraphics[width=\linewidth]{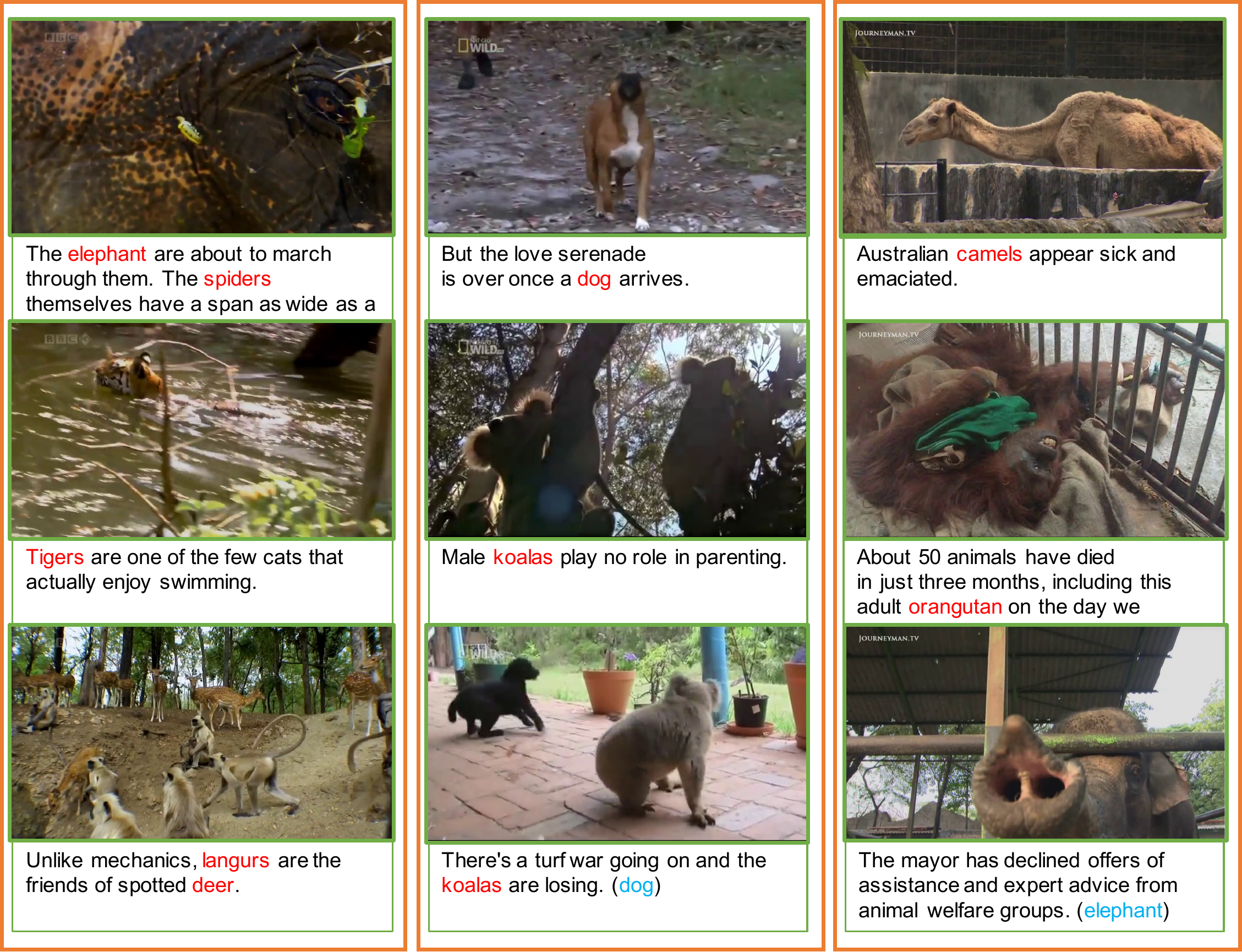}
\vskip -0.2cm
\caption{\small We wish to devise a method that can \emph{automatically} learn object detectors as it watches a documentary, 
in a weakly supervised fashion. The figures show some examples of the proposed Wildlife Documentaries (WLD) dataset. The nouns are marked in red in there is a correspondence between subtitle and object in the video. Otherwise, we provide the corresponding noun in blue color when there is a mismatch.}
\label{fig:wld-dataset}
\end{center}
\vspace{-0.8cm}
\end{figure}

These difficulties call for an alternative approach to learning object detectors.
{\em ``There is no record of an Orca doing any harm in the wild.''} -- 
we were deeply touched by this as we watched \emph{Blackfish}, 
a famous documentary film directed by Gabriela Cowperthwaite. 
For many of us, the knowledge about various creatures on the planet 
are learned from documentaries as such. 
Now, as we move steadily towards the era of AI, 
we are motivated to ask: {\em Can computers learn in a similar way?}

From an intuitive perspective, 
the idea of exploiting documentaries to learn object detectors is promising. 
Thousands of documentary films that cover a broad range of topics are produced every year, 
and the number is rising rapidly. 
A documentary film usually comprises a number of visual presentations with subtitles 
that cover the entities being introduced from different perspectives. 
Clearly, this is a rich source of information for visual learning.

%\begin{figure}[t]
%\begin{center}
%\includegraphics[width=1\linewidth]{figures/empty.png}
%\caption{Our approach is capable of learning from documentaries to generate accurate object detections of unseen classes.}
%\label{fig:overview}
%\end{center}
%\vspace{-0.5cm}
%\end{figure}

In this work, we aim to derive a method that can 
\emph{automatically} learn object detectors as it watches a documentary, 
in a weakly supervised fashion (Fig.~\ref{fig:wld-dataset}).
Towards this goal, we are facing multiple challenges.
First and foremost, the framework has completely no prior knowledge 
as to what it will see from the videos, 
\eg~the number of object categories and how the story is going to be presented. 
Everything has to be discovered from the data 
-- no human assistance is available.
Second, associations between linguistic cues and visual cues are often ambiguous. 
For example, when multiple objects are present at the same time, 
it is difficult to identify the correspondence 
between the \emph{nouns} in a subtitle and the \emph{objects} that appear in the associated frames. 
The ubiquitous pronouns and synonyms further complicate the problem. 
-- a \emph{tiger} is often called a \emph{cub} or a \emph{big cat}, 
it can also be referred to using pronouns, such as \emph{``it''} or \emph{``this''}.
Third, there are substantial variations of an object’s appearance 
due to position changes, blurring, and partial occlusions. 
Even state-of-the-art recognition techniques may experience difficulties 
in handling all such circumstances. 

The difficulties are clear -- 
whereas the materials contained in documentaries are rich and relevant, 
the observations that we work with are filled with 
noises, ambiguities, and disconnected pieces. 
In tackling this challenge, we develop a \textit{probabilistic framework}, 
with an aim to bridge individual pieces of information, 
turning them into a coherent picture. 
Specifically, the framework incorporates three kinds of factors as \emph{bridges}: 
(1) the \emph{appearance} factor that captures the common appearance pattern of each object category, 
(2) the \emph{geometric} factor that stitches isolated tracklets of an object into a whole, and 
(3) the \emph{grounding} factor that associates linguistic references with visual observations. 
In addition, the framework also identifies nouns and pronouns that describe the same entities via coreference analysis. 
On top of this formulation, we further derive an learning algorithm, 
where the learning of object detectors and the mining of training data 
are conducted along with each other within a unified optimization procedure.

%To test the performance of the proposed method, 
%we constructed a dataset that contains about $6$ hours of videos, 
%where all objects are annotated frame-by-frame (for evaluation purpose only). 
%Experimental results demonstrate that 
%this method can automatically discover new object categories 
%from videos and effectively learn detectors therefrom, 
%both without the need of human intervention.

The main \textbf{contributions} of this work lie in several aspects:
(1) a novel approach to learning object detectors, namely, 
learning from documentary videos in a weakly supervised way, without any annotated object seeding or bag-level supervision as in multiple instance learning methods.;
(2) a framework that can effectively integrate noisy pieces of information, 
including visual and linguistic cues; and
(3) a new dataset with detailed annotations (Fig.~\ref{fig:wld-dataset})\footnote{Dataset and code are available at \url{https://github.com/hellock/documentary-learning}.}.
%which not only helps assess the proposed method but also facilitates future research along this line.

%% file: related-work.tex
\section{Related Work}
\label{sec:related}
\vspace{-0.2cm}
\noindent
\textbf{Weakly-supervised object localization.} The time-consuming annotation process in object detection can be sidestepped by weakly supervised learning~\cite{bagon2010detecting,bilen2015weakly,cinbis2016weakly,deselaers2012weakly,hoffman2015detector,pandey2011scene,prest2012learning,ren2016weakly,shi2013bayesian,siva2012defence,song2014learning,song2014weakly,wang2014weakly}. In many cases, the supervised information is restricted to binary labels that indicate the absence/presence of object instances in the image, without their locations. Typically, a multiple instance learning~\cite{maron1998framework} framework is adopted. Specifically, each image is considered as a `bag' of examples given by object proposals. Positive images are assumed to contain at least one positive instance window, while negative images do not have the object at all. A good overview is provided in \cite{cinbis2016weakly}.
Prest \etal~\cite{prest2012learning} introduce an approach for learning object detectors from real-world web videos known only to contain objects of a target class. In other words, their study requires one label per video. Our problem is more challenging in that documentaries do not provide precise labels even at the `bag' level, so we do not have definite positive and negative windows in videos. We thus require an effective framework to integrate noisy pieces of information.
Joulin \etal~\cite{TangECCV14} propose a method to localize objects of the same class across a set of distinct images or videos. The co-localization problem assumes each frame in a set of videos contains one object of the same category.

Other relevant studies include \cite{liang2015towards,misra2015watch} that perform semi-supervised learning to iteratively learn and label object instances from long videos. Annotated seeds are required in these approaches.
Kumar \etal~\cite{kumar2016track} transfer tracked object boxes from weakly-labeled videos to weakly-labeled images to automatically generate pseudo ground-truth boxes. Our approach works directly on video.
Alayrac \etal~\cite{alayrac2016unsupervised} model narrated instruction videos to learn the sequence of main steps to complete a certain task. Many videos with transcript of the same
task are required.
Kwak \etal~\cite{kwak2015unsupervised} formulate the problem of unsupervised spatio-temporal object localization as a combination of discovery and tracking. It locates one instance each frame and cannot align semantic labels with clusters of objects.
Recent studies~\cite{oquab2015object,zhou2014object} show that object detectors emerge within a convolutional neural network (CNN) trained with image-level labels. We leverage on this concept to generate candidate proposals in our approach.

\noindent
\textbf{Grounding objects from image descriptions.}
A number of approaches have been proposed to localize objects in an image given its description. For instance, Karpathy \etal~\cite{karpathy2014deep} address the inter-modal alignment problem by embedding detection results
from a pretrained object detector and the dependency tree from a parser with a ranking loss.
Plummer \etal~\cite{plummer2015flickr30k} learn a joint embedding of image regions and text snippets using Canonical Correlation Analysis (CCA) to localize objects mentioned in the caption.
Recent studies~\cite{hu2016natural,mao2016generation} build upon image captioning frameworks such as LRCN~\cite{donahue2015long} or m-RNN~\cite{mao2015deep}, which are trained with ground-truth phrase-region pairs of known object classes. The idea is extended to image segmentation from natural language expressions~\cite{hu2016segmentation}.
Rohrbach \etal~\cite{rohrbach2016grounding} present a latent attention approach that learns to attend to the right region of an image by reconstructing phrases. 
In contrast to aforementioned studies, our work neither assumes objects of seen classes nor any paired ground-truth phrase-image or phrase-region data. Instead, we focus on discovering and learning to detect unknown objects with the help of unstructured linguistic references.

\noindent
\textbf{Linguistic cues in vision.}
Subtitles have been exploited for assisting the learning of visual recognizer.  Several studies \cite{buehler2009learning,cooper2009learning,pfister2014domain} automatically learn British Sign Language signs from TV broadcasts. Their videos contain a single signer with a stable pose. Positive correspondence between subtitle and signing can be easily identified due to the more structured data.  Ours contain multiple animals moving around, exhibiting various poses and scales. Our problem thus requires a more error-tolerant formulation for learning and linking the appearance, geometric, and grounding factors.
Another study~\cite{everingham2006hello} reduces ambiguity in automatic character identification by aligning subtitles with transcripts that contain spoken lines and speaker identity.  In our case, we do not have access to transcripts.
Some studies~\cite{everingham2006hello, ramanathan2014linking,bojanowski2013finding} explore transcripts of movies and TV series to help identify characters. A similar idea is proposed in~\cite{laptev2008learning} for action recognition. The character names are provided in advance, but in our setting, we assume that categories are unknown so that new objects can be discovered. %The information from textual data in our case is much weaker than Movie/TV scripts, which provide richer contexts such as speaker identities which are given next to the scene description.

%% file: dataset.tex
\section{Wildlife Documentaries (WLD) Dataset}
\label{sec:dataset}

Right from the beginning of this study we ruled out videos with relatively clean separation of objects and background for our study of unknown objects discovery. Instead, we wish to mine for more meaningful and richer information from complex videos.
To facilitate our study, we collect a new dataset called Wildlife Documentaries (WLD) dataset. It contains 15 documentary films that are downloaded from YouTube. The videos vary between 9 minutes to as long as 50 minutes, with resolution ranging from 360p to 1080p.
A unique property of this dataset is that all videos are accompanied with subtitles that are automatically generated from speech by YouTube. The subtitles are revised manually to correct obvious spelling mistakes.
To facilitate evaluations, we annotate all the animals in the videos resulting in more than \textbf{4098} object tracklets of \textbf{60} different visual concepts, \eg, `tiger', `koala', `langur', and `ostrich'.
We show some examples in Fig.~\ref{fig:wld-dataset}.

The WLD dataset differs from conventional object detection datasets in that it is mainly designed to evaluate an algorithm's capability in \textit{discovering object of unknown classes} given with rich but ambiguous visual and linguistic information in videos. The videos are much longer and are left as they are
without manual editing, while existing datasets usually provide short video
snippets. The total frame number is more than \textbf{747,000}. Object bounding box annotations are not designated for model training, but provided to evaluate how accurate an algorithm could associate the object tubes with the right visual concepts.

The dataset is challenging in two aspects. Since videos are long documentaries, large variation in scale, occlusion and background clutters is common. In many cases multiple objects co-exist in a frame. This adds difficulty in associating a target object with the correct nouns. 
Besides the visual challenges, highly unstructured subtitles also add to complexity. As can be seen from Fig.~\ref{fig:wld-dataset}, meaningful nouns are overwhelmed by abundant of pronouns and synonyms. The occurrence of a noun
does not necessarily imply the presence of the corresponding object due to temporal distance between object and subtitle. It is possible that a correspondence do not occur at all.

%% file: framework.tex
\begin{figure}[t]
\begin{center}
\includegraphics[width=\linewidth]{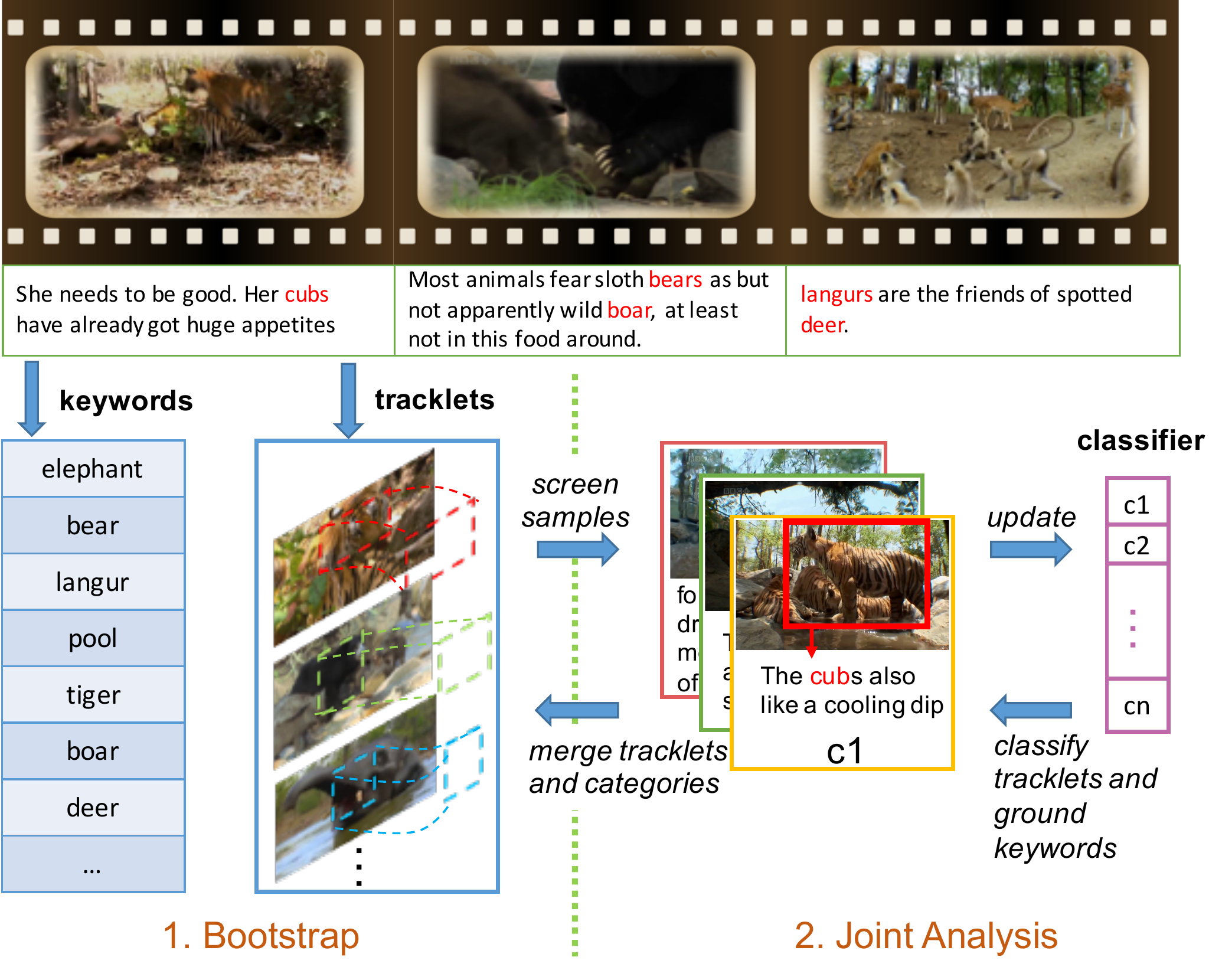}
\vskip -0.4cm
\caption{\small The proposed framework learns object detectors from documentary videos in an unsupervised manner. This is made possible through integrating noisy pieces of information, including visual and linguistic cues.}
\label{fig:frmwork}
\end{center}
\vspace{-0.7cm}
\end{figure}

\section{Framework Overview}
\label{sec:framework}

Our primary goal in this work is to develop a framework that can
\emph{discover} new objects from a documentary video, 
and \emph{learn} visual detectors therefrom. 
Note that the framework knows nothing about the 
given video a priori. 
The video itself and the associated subtitles, 
are the only data that it can rely on in the analysis.
 
As shown in Figure~\ref{fig:frmwork}, our framework accomplishes this
task through two major stages, namely 
the \emph{bootstrap} stage and the \emph{joint analysis} stage. 
The \emph{bootstrap} stage is to prepare the materials
for analysis. 
Specifically, it will acquire
a collection of \emph{tracklets} using a class-agnostic detector and tracker, 
and extract a list of \emph{keywords} via low-level linguistic analysis. 
The \emph{joint analysis} stage that ensues
aims to assemble the tracklets and keywords into a coherent and meaningful picture. 
This analysis is devised on top of a CRF formulation,
which unifies several tasks into an iterative optimization procedure.
At each iteration, it selects a subset of \emph{confident} samples
to train or fine-tune the classifier for each object category,
employs the updated classifiers to recognize the classes of individual tracklets,
grounds the keywords to them accordingly, and 
merges isolated tracklets based on the updated knowledge.
In this way, object classifiers can be gradually improved with a
growing and refined pool of training samples. 

%% file: bootstrap.tex
\section{Bootstrap}
\label{sec:bootstrap}

The bootstrap stage is to prepare the inputs for joint analysis, which includes
\emph{candidate tracklets} and a list of \emph{keywords} extracted from subtitles.

\noindent
\textbf{Obtain candidate tracklets.}
We employ an simplified version of the method proposed by Kang \etal~\cite{kang2016object} to
generate candidate tracklets, without performing the actual detection task. 
Specifically, we first use a Region Proposal Network (RPN)~\cite{ren2015faster} to generate class-agnostic 
object proposals from each video frame, and then apply a CNN-based classifier
to evaluate the \emph{objectness} score of each proposed bounding box.
Then, we extend each proposal with high score to a \emph{tracklet} via tracking.
%
%Particularly, we use a tracker~\misscite to propagate the proposal across
%frames in both forward and backward direction, until the tracker confidence
%drops below a certain threshold.
%%
%The tracklet will be further refined as follows.
%At each frame that the tracklet passes, we randomly sample bounding boxes
%around the original one, and evaluate the objectness scores for them.
%Among all these boxes, the one with highest score will be selected to indicate 
%the tracklet position at that time step.
%
Note that we re-train the RPN by excluding any object of interests in the WLD dataset to make sure our approach is agnostic to the categories.  
The CNN for evaluating objectness provides an appearance feature for each bounding box,
which will be retained and used in joint analysis.

\noindent
\textbf{Select keywords.}
Our linguistic analysis takes four steps to select keywords from subtitles:

\noindent
\textit{(1) Coreference resolution.}
Pronouns are ubiquitous in typical narratives. 
To address this, we first use the method described in~\cite{martschat2015latent} to resolve the correspondences between pronouns and nouns and substitute all pronouns with
the corresponding nouns, \eg~from \emph{``his paw''} to \emph{``tiger's paw''}. 

\noindent
\textit{(2) POS tagging.}
Object categories are usually nouns. To identify them, we apply
\emph{Part-Of-Speech (POS)} tagging using a parser~\cite{bird2009natural}. 

\noindent
\textit{(3) Lemmatization.}
In English, a word can appear in a sentence in different forms, 
\eg~\emph{``bear''}, \emph{``bears''}, and \emph{``bear's''}. 
This step is to replace different forms of a word with its canonical form,
so they will be considered as the same in later analysis.

\noindent
\textit{(4) Selection.}
Finally, we want to select a subset of nouns as \emph{keywords}
that may indicate object categories. 
To avoid common words such as \emph{``animal''} and \emph{``food''},
we resort to \emph{tf-idf scores}~\cite{salton1986introduction}, motivated by the rationale that
relevant keywords should frequently appear in only a small number of documents.
To obtain reliable \emph{tf-idf scores}, we collect a corpus that contains
$2000$ documentary transcripts and compute the scores thereon.
We found empirically that the procedure outlined above can provide
a good list of keywords that cover most of the animals appearing in 
our dataset.

%% file: model.tex
% !TEX root = ../main.tex

\section{Joint Analysis}
\label{sec:model}

Following the \emph{bootstrap} stage,
the framework will perform joint analysis based on a probabilistic model
-- 
classify each tracklet, infer the associations between tracklets and keywords, 
and as well obtain new object classifiers.

\subsection{Task Statement}

Specifically, given a video, the inputs at this stage include two parts:
(1) \textbf{Tracklets.}
We denote all tracklets as $\cT = \{\tau_1, \ldots, \tau_n\}$.
Each tracklet $\tau_i$ is a sequence of bounding boxes,
and can be described by a \emph{visual feature} $\vv_i$
and a \emph{geometric feature} $\vu_i$.
The former is formed by the appearance features extracted from a sub-sequence
of frames, as $\vv_i = (\vv_i^{(1)}, \ldots, \vv_i^{(l_i)})$;
while the latter captures the spatial/temporal characteristics of the
bounding boxes.
(2) \textbf{Keywords.}
We consider each subtitle as a \emph{bag of keywords} for simplicity.
Putting all subtitles together, we can thus obtain a large collection of keywords,
denoted by $\cW = \{w_1, \ldots, w_m\}$.
Each keyword has a time span, which is the same as that of the parent subtitle.

The purpose of this \emph{joint analysis} stage is to 
accomplish three key tasks:
(1) \textbf{Categorization.}
An important goal of this work is to detect objects from a given video.
This is accomplished by assigning a category label
$z_i \in \cC$ to each candidate tracklet $\tau_i$.
Here, $\cC$ is the set of all categories,
including all object categories and
a \emph{background} category with label $0$.
%Particularly, $z_i = 0$ indicates that the tracklet $\tau_i$ belongs to
%the background, \ie~none of the object classes.
%
(2) \textbf{Grounding.}
Along with a subtitle, multiple tracklets may appear in the scene. 
To bridge the visual and the linguistic domains,
we need to \emph{ground} the keywords to individual tracklets, 
\ie~determine which keywords correspond to which tracklets.
Generally, a keyword may be grounded to zero or multiple tracklets, and
vice versa a tracklet may be referred to by multiple keywords.
Here, we use $a_{ij} \in \{0, 1\}$ to indicate whether the tracklet $\tau_i$
is associated with the keyword $w_j$.
(3) \vspace{2pt} \textbf{Classifier Learning.}
The detected tracklets with their inferred labels constitute
a training set on which object classifiers can be learned. 
Specifically, we can select a \emph{confident} subset of tracklets 
classified to each object category and train a classifier thereon.

\subsection{Probabilistic Formulation}

In this work, we propose a Conditional Random Field (CRF) that unifies all these tasks
into a probabilistic formulation, which comprises the following potentials:

\vspace{2pt}
\noindent
\textbf{Appearance potential} $\psi_{ap}(z_i | \vv_i; \vtheta)$:
This potential measures how well a tracklet $\tau_i$ matches an object category $z_i$
based on its appearance feature $\vv_i$. It is defined as
\begin{equation}
\psi_{ap}(z_i | \vv_i, \vtheta) = \sum\nolimits_{t=1}^{l_i} \log p(z_i | \vv_i^{(t)}; \vtheta).
\end{equation}
When a convolutional network is used, $p(z | \vv; \vtheta)$ is simply
the logarithm of the output of the softmax layer, and
the parameters $\vtheta$ are the network weights.

\vspace{2pt}
\noindent
\textbf{Keyword-tracklet potential} $\phi_{kt}(z_i, a_{ij} | \veta)$:
As mentioned, $a_{ij}$ indicates whether the tracklet $\tau_i$ associates with the keyword $w_j$.
The value of this potential is determined by the object category $z_i$, as
\begin{equation}
    \phi_{kt}(z_i, a_{ij} | \veta) = \log p(z_i | w_j; \veta).
\end{equation}
Each object category can have multiple keywords, \eg~class \emph{tiger}
have keywords \emph{``tiger''} and \emph{``cub''}.
Here, $p(z_i | w_j; \veta)$ is the probability that the keyword $w_j$ belongs to the class $z_i$,
and the parameter $\veta$ is the conditional probability table.
A restriction is enforced here -- each keyword can only be grounded to
a tracklet whose time span overlaps with its own.
In other words, $a_{ij}$ is forced to be zero when the time spans of $\tau_i$ and $w_j$
have no overlap.

\vspace{2pt}
\noindent
\textbf{Geometric potential} $\phi_{st}(r_{ii'}, z_i, z_{i'} | \vu_i, \vu_{i'})$:
Here, $r_{ii'}$ indicates whether tracklets $\tau_i$ and $\tau_{i'}$
are two consecutive segments of an object trajectory and thus should be merged.
The value of $\phi_{st}$ is defined to be
\begin{equation}
    \phi_{st}
    = \begin{cases}
         \delta(z_i = z_{i'}) \cdot s(\vu_i, \vu_{i'}) & (r_{ii'} = 1), \\
         0 & (r_{ii'} = 0).
    \end{cases}
\end{equation}
Here, $s(\vu_i, \vu_{i'})$ is the spatial/temporal consistency. %Details of computation is provided in the supplementary material.
This definition ensures that two tracklets can only be merged when
they have the same class label and are consistent in both space and time.

\vspace{2pt}
\noindent
\textbf{Joint model.}
The joint CRF can then be formalized as
\begin{multline}
    p(\vz, \va, \vr | \vo; \Theta) = \frac{1}{Z(\Theta)}
    \exp \left(
        \Psi_{ap}(\vz | \vo; \vtheta) + \right. \\
        \left. \Phi_{kt}(\vz, \va | \vo; \veta) +
        \Phi_{st}(\vr, \vz | \vo)
    \right).
\end{multline}
Here, $\vz$, $\va$, and $\vr$ are the vectors that respectively comprise
all tracklet labels $(z_i)$, keyword-tracklet association indicators $(a_{ij})$,
and tracklet link indicators $(r_{ij})$.
$\vo$ denotes all observed features, and $\Theta$ are model parameters.
The three terms are given by
\begin{align}
    \Psi_{ap}(\vz | \vo; \vtheta) &= \sum\nolimits_{i=1}^n \psi_{ap}(z_i | \vv_i; \vtheta), \\
    \Phi_{kt}(\vz, \va | \vo; \veta) &= \sum\nolimits_{(i, j) \in \cG} \phi_{kt}(z_i, a_{ij} | \veta), \\
    \Phi_{st}(\vr, \vz | \vo) &= \sum\nolimits_{(i, i') \in \cR} \phi_{st}(r_{ii'}, z_i, z_{i'} | \vu_i, \vu_{i'}).
\end{align}
Here, $\cG$ is the set of tracklet-keyword pairs that can possibly be associated,
\ie~their time spans overlap;
$\cR$ is the set of all tracklet-tracklet pairs which can possibly be merged,
\ie~spatial-temporal consistency is sufficiently high.

\subsection{Joint Learning and Inference}
\label{subsec:joint_learning_inference}

Given a video, we employ \emph{variational EM} to estimate
the parameters $\vtheta$ and $\veta$, and as well infer the latent variables,
including $(z_i)$, $(a_{ij})$ and $(r_{ii'})$.

\vspace{-5pt}
\paragraph{Initialization.}
To begin with, we form the initial set of classes by considering each 
distinct keyword as a class -- such classes may be merged later 
as the algorithm proceeds. 
Also, to provide initial labels, we cluster all tracklets into
a number of small groups based on their appearance using Mean Shift.
Each cluster will be assigned a label according to 
the most frequent keyword among all those that overlap 
with the tracklets therein.
Our experiments will show this heuristic method, while simple and not very accurate, does
provide a reasonable initialization for joint analysis.

\vspace{-5pt}
\paragraph{Iterative Optimization.}
The main algorithm is an iterative procedure that alternates
between the following steps:
\begin{enumerate}[leftmargin=12pt,itemsep=0ex]

\item \textbf{Screen samples.} 
We observed detrimental impacts if all tracklets are fed to 
the training of classifiers, especially in the initial iteration, 
where the inferred labels can be very noisy. 
To tackle this issue, we explicitly enforce a screening mechanism,
a confidence value will be computed for each tracklet
based on a number of metrics, 
\eg~the length, the mean objectness score, the stability of the
objectness scores, as well as the classification margin, \ie~the difference 
between the highest classification score and the runner-up.
These metrics are combined using an SVM trained on a subset
held out for tuning design parameters.

\item \textbf{Update classifiers.}
With the confidence values, all tracklets whose confidence is beyond 
a certain threshold will be gathered to train or fine-tune the object classifiers of the corresponding categories. 
For each tracklet, we consider the class with highest score
as its ``true class'' for the current iteration.
The tracklets are sub-sampled with a fixed interval
in order to save training time.
The sampling interval is determined in a way such that
the number of samples in all classes are relatively balanced.
In addition, we also train a classifier for the \emph{background class},
so as to enhance the contrast between foreground objects and the background.

\item \textbf{Classify tracklets.}
With the updated classifier, we can infer the posterior probabilities of 
the class label $z_i$ for each tracklet $\tau_i$, denoted by $q_i$, as
{\small
\begin{equation}
    q_i(z_i) \propto 
    \exp \left(
        \psi_{ap}(z_i | \vv_i, \vtheta) + 
        \sum_{j \in \cG_i} \phi_{kt}(z_i, a_{ij} | \veta)
    \right).
\end{equation}
} 
Here, $\cG_i = \{j: (i, j) \in \cG\}$ is the set of keywords
that overlap with $\tau_i$.  
Here, the inference of the label $z_i$ considers 
not only the appearance (1st term) but also the keywords 
associated with it (2nd term).

\item \textbf{Ground keywords.}
First, each keyword $w_j$ can only be grounded to those tracklets 
with overlapping time spans. 
For one of such tracklets, whether $w_j$ should be grounded to $\tau_i$
depends only on the class label $z_i$. 
Particularly, the posterior probabilities of $a_{ij}$ of given by
$p(a_{ij} | z_i) \propto \exp( \phi_{kt}(z_i, a_{ij} | \veta) )$.
We found empirically that for most of the cases the probabilities
are close to either $0$ or $1$. 
Hence, we simply sample $a_{ij}$ therefrom to determine the grounding relations.

\item \textbf{Merge tracklets.}
For each pair of tracklets $\tau_i$ and $\tau_{i'}$, we sample
$r_{ii'}$ based on $\phi_{st}(r_{ii'}, z_i, z_{i'} | \vu_i, \vu_{i'})$.
If $r_{ii'} = 1$, they will be merged into a new single tracklet.
From the next iteration, the newly merged tracklet will be treated as a whole.
Over time, the pool of isolated tracklets will be gradually consolidated
into longer trajectories.

\item \textbf{Merge categories.}
At the end of each iteration, we compute the similarity between 
each pair of categories using the \emph{Earth Mover's Distance}~\cite{rubner2000earth},
and merge similar classes (\ie~the distance below a threshold) 
into a single one.
Along with this, the tracklet labels of the merged classes will be 
re-mapped to the new labels accordingly.

\end{enumerate}

As we can see, multiple tasks are integrated in this procedure.
Through iterative updates, a coherent picture over the given video 
will be gradually formed, where tracklets are assigned to object categories,
keywords are grounded to relevant tracklets, and more importantly,
a set of new object classifiers will be derived. 
Note that this is not a strict optimization algorithm, as two additional
steps are inserted, which include \emph{screen sampling} and \emph{category merging}.
While such steps are not directly derived from the CRF formulation, 
they do play significant roles in guiding the procedure towards a desirable direction.

%% file: experiments.tex
\section{Experiments}
\label{sec:experiments}

We evaluated our framework on the WLD dataset,
comparing the results from different iterations and 
those from a supervised counterpart.
We also studied the contributions of different components.

\noindent
\textbf{Settings.} 
Recall that we use the RPN~\cite{ren2015faster} for proposals generation and a CNN for objectness scoring and feature extraction in the bootstrap stage. The CNN is later used as a classifier in the joint analysis stage by adding fully-connected and softmax layers. For efficient iterative optimization during joint analysis, we only update the fully-connected layers and keep the convolutional layers fixed. We use ResNet-269 for the RPN and ResNet-101 for the CNN. Both models are trained using images excluding the object of interests in WLD.
 
%The RPN~\misscite and Fast R-CNN used for generating tracklet proposals are trained on a subset of ImageNet using ResNet-269 model.
%All images containing classes overlapping with WLD are excluded to avoid unexpected supervision, ensuring that the detection model has no prior knowledge of objects in the documentary videos.
%Tracking bounding box proposals in a long video is very time consuming.
%To balance the efficiency and quality, we use DSST~\cite{danelljan2014accurate} instead of FCNT~\misscite, which is much faster though less accurate.
%The classifier has exactly the same convolution layers as Fast R-CNN, followed by two fully connected layers.
%Parameters of convolution layers are fixed and initialized using the trained Fast R-CNN model, which making the training process simple and fast.

\noindent
\textbf{Evaluation metrics.} 
Since our approach is unsupervised, we evaluate our method on the whole WLD dataset and the primary metric is precision-recall.
Some video detection tasks such as ImageNet VID evaluate the performance the same way as in object detection in images, regardless of the temporal accuracy and tracklet instance matching.
On the contrary, we evaluate the results at the tracklet-level instead of the box-level, which can reflect the detection results more intuitively and precisely. Specifically, the Intersection-over-Union (IoU) criterion is extended to spatio-temporal tracklets, the 3-D IoU of tracklet $\tau_p$ and $\tau_{gt}$ is calculated as $\frac{volume(\tau_p\cap\tau_{gt})}{volume(\tau_p\cup\tau_{gt})}$.
We use $0.3$ as the threshold in all of our experiments.

\subsection{The Effectiveness of Iterative Optimization}
\label{subsec:iteration_comparison}

\begin{figure}[t]
\begin{center}
   \includegraphics[width=0.8\linewidth]{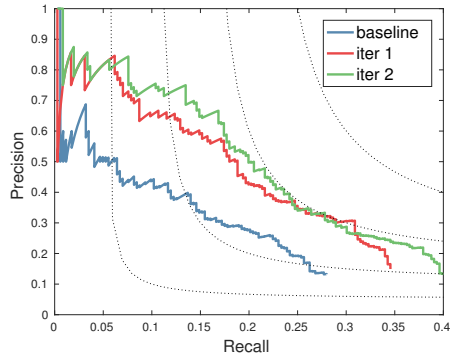}
\end{center}
\vskip -0.7cm
   \caption{\small The overall performance of different iterations.}
\label{fig:overall-pr}
\end{figure}

\begin{table}[t]
\caption{\small Average precision (\%) on a subset of WLD dataset.}
\vspace{-0.5cm}
\small{
\begin{center}
\begin{tabular}{*{4}{c}}
\hline
 & Iter 0 & Iter 1 & Iter 2 \\
\hline
mAP & 8.0 & 8.3 & 8.7 \\
discovered mAP & 20.6 & 28.9 & 30.7 \\
\hline
\end{tabular}
\end{center}
}
\label{tab:overall-ap}
\vspace{-0.5cm}
\end{table}

Our learning method converges after 2 iterations and the results are shown in Fig.~\ref{fig:overall-pr}. The results suggest the effectiveness of our joint analysis step in incorporating noisy visual and linguistic information. It is observed that more iterations only give marginal improvements. This may due to that a majority of confident samples have been
mined and utilized in the first two iterations, and little extra information can be found in further steps.
We also measure the mAP of our results like other supervised detection methods, as shown in Table~\ref{tab:overall-ap}.
Different from supervised and some weakly-supervised methods that have a prior of object categories,
candidate categories are derived from subtitle analysis in our framework,
so some categories with only a few objects, \eg~\emph{elephant}, may be hard for our framework to discover.
Consequently, we propose \emph{discovered mAP} besides the regular mAP, which means the mAP of all discovered categories.
The results shows that object categories mentioned more often in the video are more likely to be discovered and learned.

\subsection{Comparison with Supervised Method}
\label{subsec:supervised}

\begin{figure}
\begin{center}
   \includegraphics[width=0.8\linewidth]{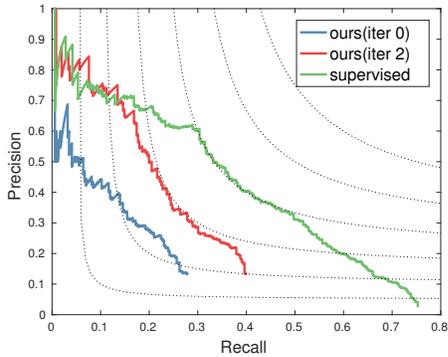}
\end{center}
\vspace{-0.75cm}
   \caption{\small Comparison with the fully supervised method.}
\label{fig:compare-supervised}
\vspace{-0.5cm}
\end{figure}

We wish to examine further how our unsupervised trained detector compares against a strong fully supervised detector.
To perform this experiment, we select categories that are available in both WLD and ImageNet. In total 7 classes are selected, which are available in a single documentary video.
Since the fully supervised detector is trained using ImageNet object detection and localization dataset,
it has prior knowledge on these selected categories, and its performance can be treated as an upper-bound of our unsupervised method.
As shown in Fig.~\ref{fig:compare-supervised}, after 2 iterations,
our unsupervised detector is competitive to the supervised counterpart in high-precision region.
The discovered mAP for the fully supervised method is $0.309$, comparable to $0.307$ of our method.
It is also observed that our method has a comparable or even higher AP in major categories of the documentary, \eg,~\emph{tiger} and \emph{langur} in this case.

% \begin{table*}
% \begin{center}
% \begin{tabular}{|*{10}{c|}}
% \hline
%  & bear & boar & deer & elephant & langur/monkey & spider & tiger & mAP & discovered mAP \\
% \hline
% Iter 0 & 24.0 & 15.3 & 5.1 & 1.3 & 11.4 & \textbf{8.3} & 22.8 & 12.4 & 12.4\\
% Iter 1 & 25.1 & 0 & 1.4 & 0 & 15.6 & 0 & 32.3 & 10.6 & 21.0\\
% Iter 2 & 23.0 & 0 & 0 & 0 & 33.1 & 0 & \textbf{36.0} & 13.2 & 30.7\\
% \hline\hline
% Supervised & \textbf{35.9} & \textbf{30.1} & \textbf{29.2} & \textbf{21.0} & \textbf{35.9} & 0 & 33.5 & \textbf{26.5} & \textbf{30.9}\\
% \hline
% \end{tabular}
% \end{center}
% \caption{The mAP (\%) of our method compared with the fully supervised one.}
% \label{tab:supervised-map}
% \end{table*}

\subsection{Ablation Study}
\label{subsec:ablation}

\begin{figure}[t]
\begin{center}
   \includegraphics[width=0.8\linewidth]{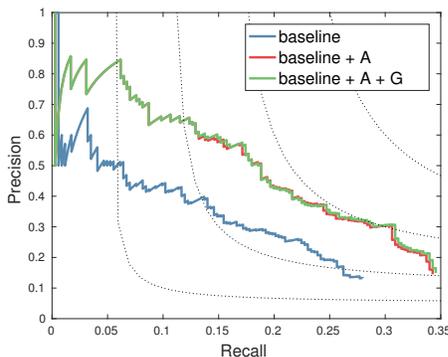}
\end{center}
\vskip -0.75cm
   \caption{\small The results of introducing different potentials. We show the results of the first iteration in joint analysis.}
\label{fig:app-geo}
\vspace{-0.2cm}
\end{figure}

To better understand how our joint model works, we study the necessity of appearance potential and geometric potential (keyword-tracklet potential is essential otherwise tracklet labels cannot be obtained).
We compare the following results: (1) Baseline, (2) Baseline + Appearance, (3) Baseline + Appearance + Geometry, as shown in Fig.~\ref{fig:app-geo}.

\noindent
\textbf{Appearance potential.}
From the results of (1) and (2), we can observe that the appearance potential contributes considerably to the performance increase.
There are two possibilities for the gain: the classifier corrected the false negative samples back to positive, or the classifier disambiguated confusing labels across different foreground categories.
To further examine the role of this potential, we treat all the foreground categories as one class and calculate the F1 score of classifier prediction compared with the baseline. We also compute the average size of top 3 largest clusters. As shown in Table~\ref{tab:cls-compare}, despite the classifier cannot recall more foreground samples compared with the baseline (reflected by the F1-binary metric),
it shows great improvement on grounding results (reflected by the F1-multiple metric).
%, which is sensitive to the cluster of samples.
When we consider the improved F1-multiple metric and the increase of cluster size, the observation suggests that the appearance model captures meaningful visual similarity and strengthen some dominant categories.
%Thus we can conclude that the classifier corrects many mistakes that confuse two tracklets other than foreground and background.

\begin{table}[t]
\caption{\small The results before and after `classify tracklets' and `ground keywords' steps (see Sec.\ref{subsec:joint_learning_inference} for details).}
\vspace{-0.5cm}
\small
\begin{center}
\begin{tabular}{*{4}{c}}
\hline
 & F1-binary & F1-multiple & average top-3 size \\
\hline
before & 0.409 & 0.240 & 96 \\
after & 0.367 & 0.305 & 277 \\
\hline
\end{tabular}
\end{center}
\label{tab:cls-compare}
\vspace{-0.5cm}
\end{table}

\noindent
\textbf{Keyword-tracklet potential.}
In our grounding method, only tracklets that are predicted as positive can be associated with keywords and all negative ones are ignored.
The positive tracklets can be noisy since the predicted labels may not be accurate. A good grounding strategy will associate correct words with true positive samples, as well as assign no word to false positive samples.
As a baseline for grounding, we employ a straightforward approach which assigns category labels based on word frequencies. 
We compare the accuracy on true positive samples of two methods using class labels of different iterations as well as ground truth labels.
The results are shown in Table~\ref{tab:grounding-accu}, 
which clearly shows the significant contribution of this potential.

\begin{table}[t]
\caption{A comparison of our grounding approach with word counting baseline accuracy.}
\vspace{-0.5cm}
\small
\begin{center}
\begin{tabular}{*{5}{c}}
\hline
 & Iter 0 & Iter 1 & Iter 2 & Ground truth \\
\hline
Word counting & 0.414 & 0.776 & 0.855 & 0.744 \\
Ours & 0.343 & 0.888 & 0.937 & 0.935 \\
\hline
\end{tabular}
\end{center}
\label{tab:grounding-accu}
\vspace{-0.7cm}
\end{table}

\begin{figure}
\begin{center}
   \includegraphics[width=0.8\linewidth]{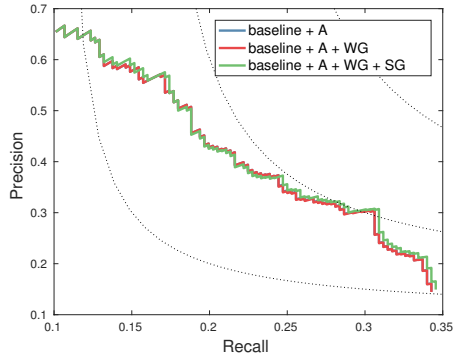}
\end{center}
\vspace{-0.7cm}
   \caption{The results of weak and strong geometric link. ``A'', ``WG'' and ``SG'' refer to ``appearance'', ``weak link'' and ``strong link'' respectively.}
\label{fig:geo-link}
%\vspace{-0.4cm}
\end{figure}

\noindent
\textbf{Geometric potential.}
Comparing (2) and (3), geometric potential also shows its effectiveness, although much weaker than the appearance potential.
The effects of geometric link potential are of two aspects named ``weak link'' and ``strong link'', which are decided by the potential value.
``Weak link'' refers to two tracklets (belonging to the same category) with weak geometric relation based on some threshold. 
``Strong link'', on the other hand, refers to two tracklets not only belong to the same category, but also strong geometric relation (which will be merged together).
We compare the following variants:
without geometric link, with weak geometric link, with both weak and strong geometric link, results are shown in Figure~\ref{fig:geo-link}.
``Weak link'' contributes little compared with ``strong link'' (completely overlapped with baseline), suggesting that the merging strategy with strong link is the most effective part in geometric potential.

\subsection{Examples and Failure Cases}

\begin{figure}
\begin{center}
  \includegraphics[width=\linewidth]{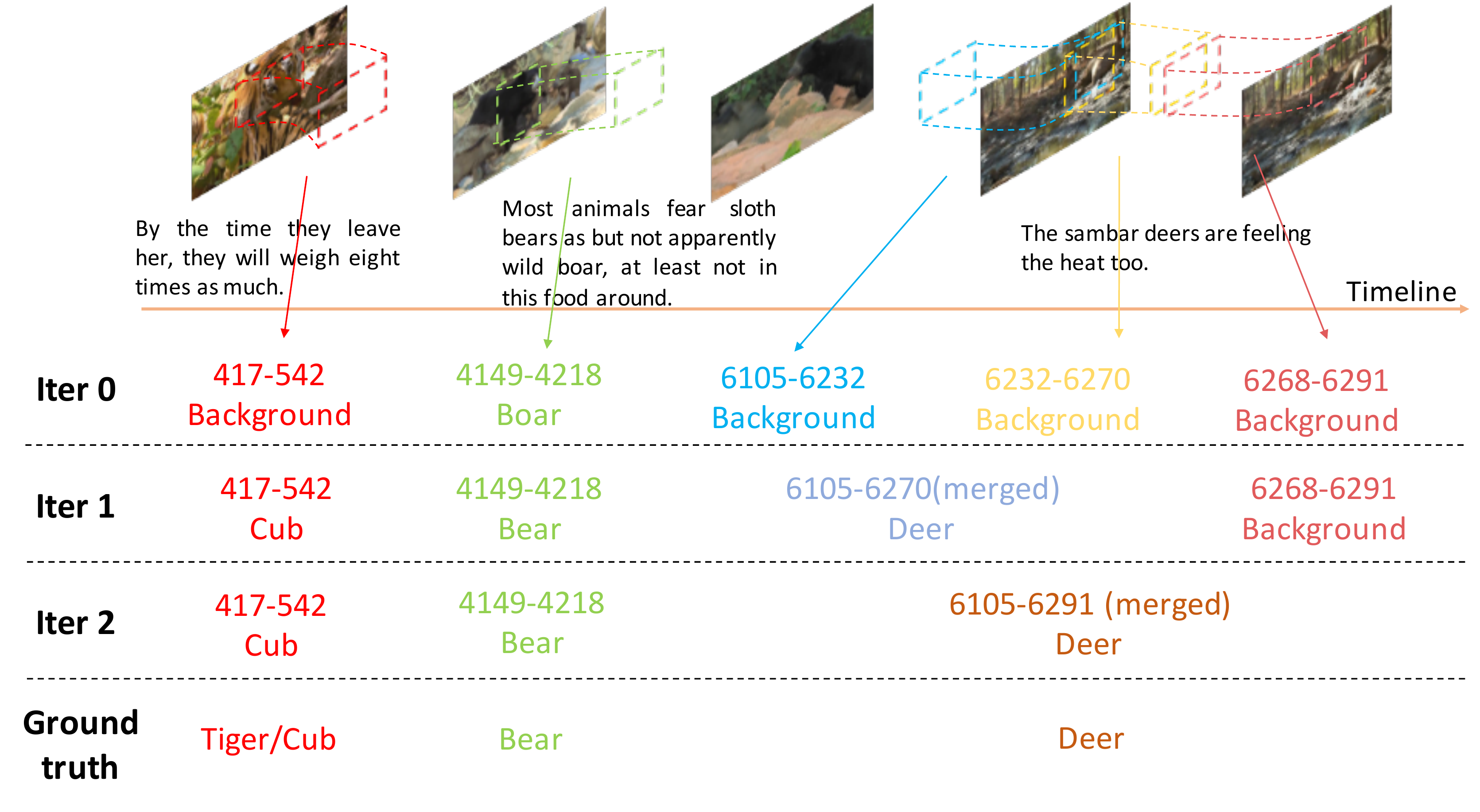}
\end{center}
\vspace{-0.5cm}
  \caption{\small Tracklet examples in different iterations.}
\label{fig:case-study}
\vspace{-0.2cm}
\end{figure}

We demonstrate some examples in Fig.~\ref{fig:case-study}, showing the results in different processes.
The initialization (iter 0) is not satisfying for the some wrong labels,
and the long trajectory is split into three short tracklet proposals.
As the iteration proceeds, the learned appearance model classifies the tracklets into correct groups,
and the strong geometric link merges separated tracklets into longer ones,
which helps to obtain the correct labels in turn.

\begin{figure}
\begin{center}
   \includegraphics[width=\columnwidth]{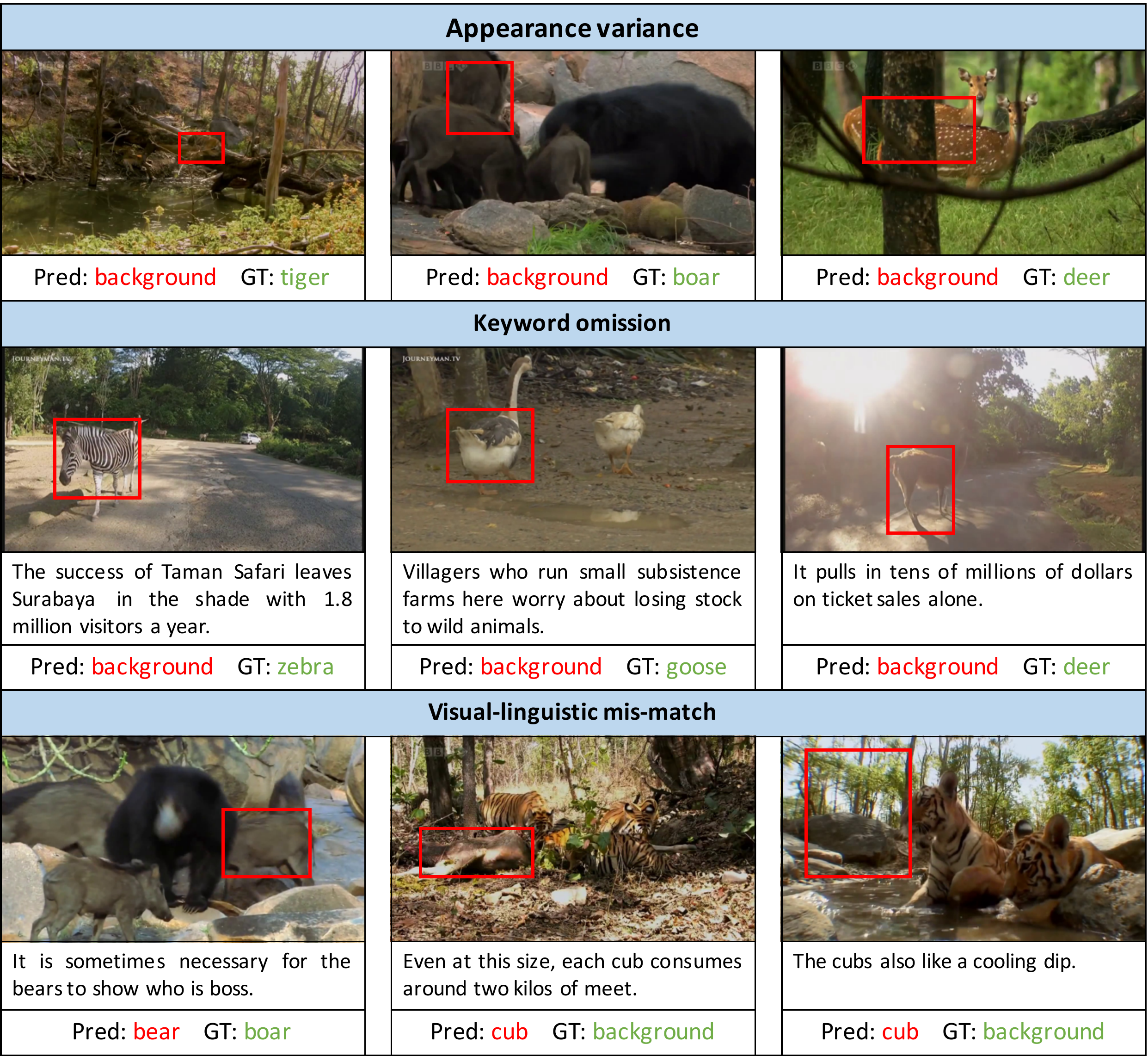}
\end{center}
\vskip -0.65cm
   \caption{Examples of failure cases.}
\label{fig:failure-cases}
\end{figure}

We also show some failure cases of our method. Figure~\ref{fig:failure-cases} illustrates three typical types of failure:
(1) contributed by the very ambiguous and challenging visual appearance of the object.
These objects are either too small or heavily occluded by other object.
(2) caused by the omission of keywords, \ie,~if the \textit{tf-idf} score is below the threshold, the category is not included in the keyword list.
(3) caused by mismatch in visual and linguistic evidences.
Specifically, if two categories usually appear at the same time or the subtitles and frames are not synchronized, it is likely to be mis-matched.

%% file: conclusion.tex
\section{Conclusion}
\label{sec:conclusion}

This paper presented a framework that can jointly discover
object categories, detect object tracks, and learn object classifiers
from documentary videos in an unsupervised fashion.
Experimental results obtained on a real world datasets
have demonstrated the effectiveness of this framework --
most animal categories are discovered automatically.
Also, as the joint analysis algorithm proceeds,
the object classifiers will be gradually improved with a
growing training set, raising the mAP from
8.0\% to 8.7\% and the discovered mAP from 20.6\% to 30.7\%.
Our framework also present comparable results with fully supervised learning for the major categories of documentaries.

\paragraph{Acknowledgment}

This work is partially supported by
the Early Career Scheme (ECS) grant (No. 24204215), and
the collaboration grant from SenseTime Group.
We would like to thank Dian Chen, Lingzhi Li, Miao Fan, Pik Kei Tse, Tsz Ching Fan, Yule Li, Zhehao Jiang for contributing to the annotation efforts.